\pdfoutput=1

\documentclass[11pt]{article}

\usepackage[]{acl}

\usepackage{times}
\usepackage{latexsym}

\usepackage[T1]{fontenc}

\usepackage[utf8]{inputenc}

\usepackage{microtype}

\title{Instructions for *ACL Proceedings}

\usepackage{times}
\usepackage{pbox}
\usepackage{latexsym}
\usepackage{cjhebrew}
\usepackage{appendix}
\usepackage{etoolbox}

\usepackage{float}
\usepackage{microtype}
\usepackage{graphicx}
\usepackage{makecell}
\usepackage{subfigure}
\usepackage{booktabs} %
\usepackage{multirow}
\usepackage{microtype}
\usepackage{cjhebrew}

\title{Breaking Character: Are Subwords Good Enough for MRLs After All?}

\author{Omri Keren$^{\dagger}$ \quad Tal Avinari$^{\dagger}$ \quad Reut Tsarfaty$^{\ddagger}$ \quad Omer Levy$^{\dagger}$  \\
\\
$^{\dagger}$ Tel Aviv University \\
$^{\ddagger}$  Bar Ilan University \\
\texttt{omrikeren@mail.tau.ac.il}}

\date{}

\begin{document}
\maketitle

\begin{abstract}
Large pretrained language models (PLMs) typically tokenize the input string into contiguous subwords before any pretraining or inference. 
However, previous studies have claimed that this form of subword tokenization is inadequate for processing morphologically-rich languages (MRLs). 
We revisit this hypothesis by pretraining a BERT-style masked language model over {\em character} sequences instead of word-pieces.
We compare the resulting model, dubbed \textit{TavBERT}, against contemporary PLMs based on subwords for three highly complex and ambiguous MRLs (Hebrew, Turkish, and Arabic), testing them on both morphological and semantic tasks.
Our results show, for all tested languages, that while TavBERT obtains mild improvements on surface-level tasks à la POS tagging and full morphological disambiguation, subword-based PLMs achieve significantly higher performance on semantic tasks, such as named entity recognition and extractive question answering.
These results showcase and (re)confirm the potential of subword tokenization as a reasonable modeling assumption for many languages, including MRLs.

\end{abstract}
\section{Introduction}

Large pretrained language models (PLMs) typically operate over contiguous subword tokens (aka word-pieces), which are created by shallow statistical methods \citep{sennrich-etal-2016-neural,kudo-richardson-2018-sentencepiece}, and do not necessarily reflect the morphological structure of  words.
This is particularly true when dealing with languages that exhibit non-concatenative morphology, such as root and pattern morphology (as in Arabic and Hebrew) or vowel harmony (e.g. Turkish).
Hence, it has been hypothesized that such  subword tokenization methods may undermine the performance of PLMs on morphologically-rich languages (MRLs) \citep{klein-tsarfaty-2020-getting,tsarfaty-etal-2020-spmrl},
with a significant body of MRL literature advocating for linguistically-informed methods, such as explicitly injecting morphological lattices into models \citep{more-etal-2018-conll, seker-tsarfaty-2020-pointer}.

\begin{figure}[t]
\includegraphics[width=0.8\columnwidth]{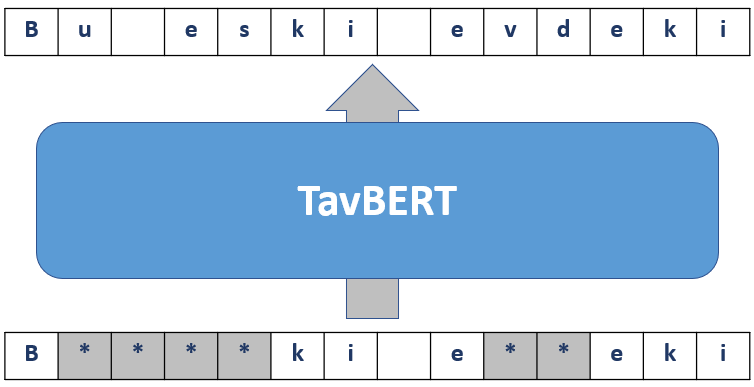}
\centering
\caption{We pretrain TavBERT by recovering randomly masked spans in the original character sequence. In this Turkish example, the tokens in asterisk are the masked characters. Whitespaces are equivalent to any other character.}
\label{fig:tavbert}
\end{figure}

In this work, we revisit the hypothesis that shallow subword tokenization is inadequate for MRLs by comparing it to a more flexible, character-aware alternative.
To that end, we train a masked language model (MLM) based on \textit{character} tokenization, TavBERT.\footnote{The word \textit{tav} refers to the word \cjRL{tw}, meaning \textit{character}, and to the last letter in the Hebrew alphabet (\cjRL{t}).}
During pretraining, we mask random spans of characters that the model then needs to predict, in a similar fashion to SpanBERT \citep{joshi-etal-2020-spanbert}.
By operating over characters rather than subwords, TavBERT has the potential to learn intricate morphological patterns that are prevalent in MRLs.

We compare TavBERT to contemporary BERT-style models trained over subword tokens \citep{antoun-etal-2020-arabert, stefan_schweter_2020_3770924, chriqui2021hebert, seker2021alephbert}, in three MRLs known to be morphologically rich and complex: Hebrew (\textit{he}), Turkish (\textit{tr}), and Arabic (\textit{ar}), on a variety of  morpho-syntactic and semantic tasks.
Experiments show that TavBERT performs on par with subword-based PLMs on part-of-speech tagging and gains only a slight advantage on full morphological disambiguation.
This indicates that subword tokenization \textit{does not} severely undermine the ability of pretrained language models to acquire morphological information, even though it obfuscates the original character sequence.
Conversely, we find that PLMs based on subword tokens significantly outperform our character-based method on the more semantic tasks in our set, named entity recognition and question answering, across all tested languages, asserting the semantic capabilities of subword-based PLMs.
Overall, our results provide evidence that, contrary to previous claims, pretraining over subword tokens constitutes a sensible inductive bias for the development of PLMs for MRLs.\footnote{Our code is publicly available at \url{https://github.com/omrikeren/TavBERT}}

\section{Model}

We aim to learn the meaningful character representations and patterns from raw text during pretraining. 
To that end, we train a masked language model (MLM) \cite{devlin-etal-2019-bert} based on the transformer encoder architecture \citep{vaswani2017attention}.
We follow SpanBERT \citep{joshi-etal-2020-spanbert} and
mask random spans of  {\em characters} for the model to predict.
We hypothesize that masking \textit{spans of characters} incentivizes the model to contextualize over longer character sequences, and to detect useful patterns.
Specifically, we sample a random starting position uniformly from the given sequence, and then sample the length of the masked span from a Poisson distribution with a parameter $\lambda$.
Each character in the span is replaced by a special \texttt{[MASK]} token.
This process is repeated until 15\% of the given sequence is masked.
Finally, the model predicts a distribution for each \texttt{[MASK]} token, which is used to compute the cross-entropy loss.
We train using the MLM objective alone, 
without the next sentence prediction (NSP) loss.
Figure~\ref{fig:tavbert} illustrates the pretraining process.

\section{Experiments}
\label{sec:experiments}
In order to test the efficacy of the character-based architecture we proposed and contrast it with standard subword-based language models for MRLs,
we experiment with two morpho-syntactic tasks, POS tagging and full morphological disambiguation, and two semantic tasks, named entity recognition (NER) and extractive question answering.

\subsection{Setup}

\paragraph{Baselines}
For all languages ({\em he/tr/ar}), we 
test multilingual BERT (mBERT) \citep{devlin-etal-2019-bert}, as well as several recently-released monolingual BERT models in their respective languages: HeBERT \citep{chriqui2021hebert} and AlephBERT \citep{seker2021alephbert} for Hebrew,
BERTurk \citep{stefan_schweter_2020_3770924} for Turkish, and
AraBERT (v0.1) \citep{antoun-etal-2020-arabert} for Arabic.\footnote{As opposed to other variants of AraBERT, v0.1 does not require a segmentation step of the raw input text.} All baseline models use BPE tokens as their underlying subwords.

\paragraph{Corpora}
We use the freely available OSCAR corpus \citep{ortiz-suarez-etal-2020-monolingual}, for pretraining (separate) TavBERT models on unlabeled text in Hebrew, Turkish, and Arabic.
Table~\ref{table:oscar-stats} details the size of the pretraining corpora for each language.

\paragraph{Vocabulary}
TavBERT's vocabulary is set to contain the top-k frequent characters whose cumulative frequency accounts for about 99.93\% of the corpus. 
Appendix~\ref{apx:vocab-stats} lists the distributions of various scripts within each language's vocabulary, and a comparison of vocabulary sizes for all tested models.

\begin{table}[t]
\small
\centering
\begin{tabular}{@{}lrr@{}} 
\toprule  
 \textbf{Language} & \textbf{File Size} & \textbf{Words} \\  
\midrule
he & 9.8G & 1.0B \\
tr & 27G & 3.3B \\
ar & 32G & 3.1B \\
\bottomrule
 \end{tabular}
 \caption[!t]{\label{table:oscar-stats} Data statistics for the pretraining set. The statistics refer to the deduplicated version of the OSCAR corpus \cite{ortiz-suarez-etal-2020-monolingual}.}
\end{table}

\paragraph{Hyperparameters}
We use Fairseq's \citep{ott-etal-2019-fairseq} implementation of RoBERTa \citep{liu2019roberta} for pretraining TavBERT models, following the \textit{base} model architecture (12 transformer encoder layers).\footnote{We set $\lambda = 5$ in our experiments to simulate the average length of BPE tokens. We do not finetune this hyperparameter.}  %
Appendix~\ref{apx:hyperparameters} details the fine-tuning hyperparameters.

\subsection{Input/Output Formats}
While TavBERT is pretrained to produce a prediction for each character, standard POS tagging and morphological disambiguation datasets, such as Universal Dependencies (UD)  \citep{nivre-etal-2020-universal}, provide labeled data over \textit{morphemes},\footnote{In UD terms, these are called {\em syntactic words}. In previous literature on Hebrew and Arabic, these are sometimes called {\em morphological segments} or simply {\em segments}.} linguistic units smaller than words. This introduces mismatches in both fine-tuning and evaluation.

We consider two mappings between morphemes and characters during fine-tuning: \textit{multitags}, and \textit{segments}. 
In the {\em multitag} variant, we simply collect all labels for the characters in each raw, space-delimited token, and assign each character of the raw token the resulting multi-set. 
In the {\em segments} variant, we assign each character the label of its encompassing morpheme. 
Appendix~\ref{apx:fine_tuning} details and illustrates each of these mapping procedures.

At inference time, 
we experiment with three heuristics for converting every model's output (i.e. character-level tags) to word-level multitags.

\paragraph{First}
The label of the first token of each word determines that of the entire word.
This heuristic is commonly used by subword models \cite{devlin-etal-2019-bert} through the canonical implementation in HuggingFace Transformers \cite{wolf-etal-2020-transformers}.

\paragraph{Majority}
The label is determined by a vote among the characters' labels, which is particularly suitable for aggregating character-level multitags.

\paragraph{Spans}
Given a word's character-level labels, we mark the maximal spans that start and end with the same label, ignoring labels in the middle of the span, %
and take the union of all the maximal spans' labels to produce the word's multitag.
For example, given the  sequence \texttt{(DET, NN, NN, VB, NN)}, we extract two maximal spans, \texttt{DET} (the first token) and \texttt{NN} (the second to fifth token), and aggregate them to produce \texttt{DET+NN}.

\subsection{Morpho-Syntactic Tasks}

To test the morphological capabilities of the models, we evaluate them on POS tagging and morphological disambiguation benchmarks.
Labeled data for both tasks is available through the Hebrew (he\_htb), Turkish (tr\_imst), and Arabic (ar\_padt) treebanks of the Universal Dependencies v2.2 dataset from the CoNLL-18 UD Shared task \cite{sade-etal-2018-hebrew}.

\paragraph{POS Tagging}
We fine-tune a token-classification head on top of the final encoder layer of each model to predict parts of speech \cite{devlin-etal-2019-bert}.
Performance is measured using the aligned multiset metric (mset-$F_1$) proposed by \citet{seker-tsarfaty-2020-pointer}, which compares the predicted word-level multitag with the ground truth's.
Table \ref{table:pos} shows that 
BPE-based BERT models do well on POS tagging in all three languages, reaching almost the same performance as TavBERT's.
These results indicate that both character- and subword-based MLMs can learn enough morphology from raw text to infer parts of speech at the morpheme level.

An error analysis for Hebrew TavBERT, performed on 50 randomly-sampled erroneous predictions from the development set, reveals that annotation inconsistencies and truly ambiguous cases account for the majority of our model's errors.
Along with our main results, these findings strongly suggest that TavBERT and other BERT-style models can reach similar agreement levels as expert human annotators, effectively solving these datasets. %

\begin{table}[t]
\small
\centering
\begin{tabular}{@{}llllr@{}} 
\toprule   
\textbf{Lang} & \textbf{Model} & \textbf{Fine-tuning}     & \textbf{Inference}    & \textbf{F1} \\  
\midrule
\multirow{7}{*}[-2pt]{he} 
& mBERT & Multitag & First & 95.25 \\
& HeBERT & Multitag & First & 96.86 \\
& AlephBERT & Multitag & First & 96.94 \\
\cmidrule{2-5}
& \multirow{2}{*}{TavBERT} %
                                 & Multitag & Majority & 96.93 \\ 
& & Segments & Spans & \textbf{97.15} \\ 
\midrule
\multirow{5}{*}[-2pt]{tr} 
& mBERT & Multitag & First & 94.55 \\
& BERTurk & Multitag & First & 96.41 \\
\cmidrule{2-5}
& \multirow{2}{*}{TavBERT} & Multitag 
& Majority & 96.50 \\ 
& & Segments & Spans & \textbf{96.61} \\ 
\midrule
\multirow{4}{*}[-2pt]{ar}
& mBERT & Multitag & First & 96.35 \\
& AraBERT & Multitag & First & 96.27 \\
\cmidrule{2-5}
& \multirow{2}{*}{TavBERT} & Multitag 
& Majority &  96.59 \\ 
& & Segments & Spans & \textbf{96.81} \\ 
\bottomrule
\end{tabular}
\caption{POS tagging results on the UD corpus in Hebrew, Turkish, and Arabic. Performance is measured by comparing word-level multitag sets (mset-$F_1$).}
\label{table:pos}
\end{table}

\paragraph{Morphological Disambiguation}
We also fine-tune the models to predict morphological features  (gender, number, person, etc.) available in each of the three languages.
In this setting, we introduce a separate token-classification head for each feature, as well as an additional head for POS tagging. All classification heads are trained jointly during fine-tuning by summing over the cross-entropy losses.
For Hebrew, in addition to UD, we fine-tune on the Hebrew section of the SPMRL shared task \citep{seddah-etal-2013-overview}.
Performance is once again measured by comparing multitags via the aligned multiset $F_1$ metric, and reported separately for POS tags and morphological features \cite{seker2021alephbert}.

Tables~\ref{table:pos-morph} and \ref{table:he-pos-morph} show the results. We observe that overall, TavBERT's performance is on par with the subword-based models, with a marginal advantage in Arabic and Hebrew. 
In terms of error reduction, TavBERT outperforms mBERT by 25\% on SPMRL and by 39\% on UD. It also surpasses the monolingual subword-based BERT models, though by a much smaller margin, namely by 10\% and 18\% error reduction relative to AlephBERT and HeBERT, respectively.

\begin{table}[t]
\small
\centering
\begin{tabular}{@{}llllr@{}} 
\toprule  
 \textbf{Lang} & \textbf{Model} & \textbf{Fine-tuning} & \textbf{Inference} & \textbf{UD} \\  
\midrule
\multirow{4}{*}[-2pt]{tr} 
& mBERT & Multitag & First & 94.98  \\
& BERTurk & Multitag & First & \textbf{96.95} \\
\cmidrule{2-5}
& TavBERT & Segments & Spans & 96.81 \\ 
& TavBERT & Multitag & Majority & 96.92 \\
\midrule
\multirow{4}{*}[-2pt]{ar} 
& mBERT & Multitag & First & 95.09 \\
& AraBERT & Multitag & First & 96.07 \\
\cmidrule{2-5}
& TavBERT & Segments & Spans & 96.42 \\ 
& TavBERT & Multitag & Majority & \textbf{97.30} \\
\bottomrule
 \end{tabular}
 \caption[!t]{\label{table:pos-morph} Aligned MultiSet (mset-$F_1$) results for morphological features on the UD corpus in Turkish and Arabic.}
\end{table}

\begin{table}[t]
\small
\centering
\begin{tabular}{@{}l@{~~~~}l@{~~~~}l@{~~~~}l@{~~~~}r@{}} 
\toprule  
 \textbf{Model} & \textbf{Fine-tuning} & \textbf{Inference} & \textbf{UD} & \textbf{SPMRL} \\  
\midrule
mBERT & Multitag & First & 94.42 & 93.72 \\
HeBERT & Multitag & First & 95.73 & 94.66  \\
AlephBERT & Multitag & First & 95.86 & 94.82 \\
\midrule
TavBERT & Segments & Spans & 96.40 & \textbf{95.33} \\
TavBERT & Multitag & Majority & \textbf{96.61} & 95.30  \\ 
\bottomrule
 \end{tabular}
 \caption[!t]{\label{table:he-pos-morph}Aligned MultiSet (mset-$F_1$) results for morphological features on the Hebrew sections of the SPMRL and UD Corpus.}
\end{table}

\subsection{Semantic Tasks}

We compare TavBERT with subword-based PLMs on extractive question answering (QA), and on named-entity recognition (NER), a task sensitive to both morphological and semantic information.

\paragraph{NER}
We use the NEMO dataset \cite{10.1162/tacl_a_00404}
for Hebrew,
the TWNERTC dataset\footnote{With the splits from \citet{rahimi-etal-2019-massively}} \cite{sahin2017automatically} for Turkish, and
the ANERCorp corpus\footnote{With the splits from \citet{obeid-etal-2020-camel}} \cite{benajiba2007anersys} for Arabic.
All three datasets provide labeled sentences at the \textit{word} level.\footnote{\citet{10.1162/tacl_a_00404} additionally propose a more granular \textit{morpheme}-based alternative.}
Performance is measured by computing the word-level $F_1$ scores on the detected entity mentions.

\begin{table}[t]
\small
\centering
\begin{tabular}{@{}llcc@{}} 
\toprule  
    \textbf{Lang} &     \textbf{Model} &    \textbf{QA}                &   \textbf{NER} \\  
                  &                    & \textsc{F1} / \textsc{EM}     & \textsc{F1} \\
\midrule
\multirow{4}{*}[-2pt]{he} 
& mBERT & \textbf{56.1} / \textbf{32.0}  & 79.07 \\
& HeBERT & 36.7 / 18.2  & 81.48  \\
& AlephBERT & 49.6 / 26.0  &  \textbf{84.91} \\
\cmidrule{2-4}
& TavBERT & 48.7 / 29.1  &  81.54  \\
\midrule
\multirow{3}{*}[-2pt]{tr} 
& mBERT &  76.6 / 56.8 & 93.53  \\
& BERTurk & \textbf{78.2} / \textbf{61.1}  & \textbf{93.57} \\
\cmidrule{2-4}
& TavBERT & 61.7 / 46.7 & 91.19 \\ 
\midrule
\multirow{3}{*}[-2pt]{ar} 
& mBERT & 81.5 / 67.1 & 
77.70
\\
& AraBERT & \textbf{83.5} / \textbf{71.1} & 
\textbf{83.48}
\\
\cmidrule{2-4}
& TavBERT & 60.0 / 45.9 & 79.45 \\ 
\bottomrule
 \end{tabular}
 \caption[!t]{\label{table:nerHeb} Results for semantic tasks. Baseline performance of NER for Hebrew is as reported by \citet{seker2021alephbert}. QA results are reported on the respective development sets, except for Hebrew, where they are reported on the test set.}
\end{table}

\paragraph{QA}
For Hebrew, we use the ParaShoot dataset \cite{keren-levy-2021-parashoot}, which contains annotated questions and answers on paragraphs curated from Hebrew Wikipedia.
For Arabic, we evaluate on all the examples in Arabic from the multilingual TyDi QA secondary Gold Passage (GoldP) task dataset \cite{tydiqa}.
For Turkish, we the TQuAD dataset\footnote{\url{https://tquad.github.io/turkish-nlp-qa-dataset/}}, which contains data on Turkish and Islamic science history.
We compare the models' predictions to the annotated answer using token-wise $F_1$ score and exact match (EM), as defined by \citet{rajpurkar-etal-2016-squad}.

\paragraph{Results}
Table \ref{table:nerHeb} shows the evaluation results on the semantic tasks. We observe that the performance gap in favor of subword models increases with the level of semantic understanding a task requires. 
Indeed, this gap is most pronounced in QA, where we observe a significant degradation in TavBERT's performance compared to subword models in all three languages.

\section{Conclusion}

This work re-examines the efficacy of subword tokenization, commonly used by pretrained language models, in morphologically rich languages. For this purpose, we introduce TavBERT, a masked language model pretrained over character spans, and compare its performance on morpho-syntactic and semantic tasks to that of contemporary BERT-style models that use BPE tokenization. Our experiments on POS tagging and morphological disambiguation for three MRLs indicate that both subword- and character-based models perform on par on morphology.
TavBERT's relatively poor performance on named entity recognition and question answering in particular, across all tested languages, suggests that models pretrained over subword tokens enjoy decent semantic capabilities, thereby serving as an appropriate modeling assumption for MRLs.

\bibliography{anthology, custom}
\bibliographystyle{acl_natbib}

\clearpage

\appendix
\section{TavBERT's Vocabulary Statistics}
\label{apx:vocab-stats}

\begin{table}[ht!]
\small
\centering
\begin{tabular}{@{}llr@{}} 
\toprule  
 \textbf{Language} & \textbf{Model} & \textbf{Vocab Size} \\  
\midrule
\multirow{3}{*}[-2pt]{he} 
 & HeBERT  & 30K \\
 & AlephBERT  & 52K \\
\cmidrule{2-3}
    & TavBERT  & 345 \\
\midrule
\multirow{2}{*}[-2pt]{tr} 
 & BERTurk & 32K \\
   & TavBERT  & 250 \\
\midrule
\multirow{2}{*}[-2pt]{ar} 
 & AraBERT (v0.1) & 64K \\
 & TavBERT & 302 \\
\bottomrule
 \end{tabular}
 \caption[!t]{\label{table:voacb-stats} Models' vocabulary sizes.}
\end{table}

\begin{table}[ht!]
\centering
\small
\begin{tabular}{@{}llc@{}} 
\toprule   
\textbf{Language} &  \textbf{Script} & \textbf{Percentage}  \\
\midrule
\multirow{4}{*}[-2pt]{he} 
   & Latin & 22\% \\
   & Cyrillic & 20\% \\
   & Hebrew & 11\% \\
   & Arabic & 7\% \\
\midrule
\multirow{2}{*}[-2pt]{tr} 
 &  Latin & 49\% \\
   & Cyrillic & 8\% \\
\midrule
\multirow{3}{*}[-2pt]{ar}
 & Arabic & 31\% \\
  & Latin & 26\% \\
   & Cyrillic & 7\% \\
\bottomrule
\end{tabular}
\caption{TavBERT vocabulary character distribution for the most common scripts, calculated out of the non-void characters.}
\label{table:tavbert-vocab}
\end{table}

\section{Hyperparameters}
\label{apx:hyperparameters}

\subsection{Pretraining}
\begin{table}[h!]
\centering
\small
\begin{tabular}{@{}lr@{}} 
\toprule   
\textbf{Hyperparameter} & \textbf{Value}  \\
\midrule
Model dimensions & 768 \\
Hidden dimensions & 3072 \\
Attention heads per layer & 12 \\
Maximal sequence length & 2048 \\
Batch size & 768 \\ 
Training steps & 125000\\
Peak learning rate & $3\mathrm{e}{-4}$\\
Warmup steps & 5000\\
\bottomrule
\end{tabular}
\caption{Hyperparamerter settings for pretraining.}
\label{table:hyperparams-pretrain}
\end{table}

\subsection{POS tagging and Morphological Analysis}
For all three languages, we select the best model by validation-set performance over the following hyperparameter grid: learning rate $\in \{3\mathrm{e}{-5}, 5\mathrm{e}{-5}, 1\mathrm{e}{-4}\}$, batch size $\in \{16, 32, 64\}$, and number of epochs $\in \{5, 6\}$.

\subsection{Named Entity Recognition}
For Hebrew, we follow the fine-tuning setting as in \citet{seker2021alephbert}.
For Turkish, we run with learning rate $5\mathrm{e}{-5}$, batch size 16, for 10 epochs.
For Arabic, we select the best model by validation set performance over the following hyperparameter grid: learning rate $\in \{3\mathrm{e}{-5}, 5\mathrm{e}{-5}, 1\mathrm{e}{-4}\}$, batch size $\in \{16, 32, 64\}$, and number of epochs $\in \{5, 6\}$, with a maximal sequence length of 320 for mBERT and AraBERT, and 2048 for TavBERT.

\subsection{Question Answering}
For Hebrew, we select the best model by validation set performance over the following hyperparameter grid: learning rate $\in \{3\mathrm{e}{-5}, 5\mathrm{e}{-5}, 1\mathrm{e}{-4}\}$, batch size $\in \{16, 32, 64\}$, and update steps $\in \{512, 800, 1024\}$. 

For Turkish, we run a sweep over the following hyperparameter grid: learning rate $\in \{3\mathrm{e}{-5}, 5\mathrm{e}{-5}, 1\mathrm{e}{-4}\}$, batch size $\in \{16, 32, 64\}$, and number of epochs $\in \{5, 6\}$.

For Arabic, we run with learning rate $3\mathrm{e}{-5}$, batch size 24, maximal sequence length 384 (1536 for TavBERT), for 2 epochs.
\clearpage

\section{Morpheme to Character Mappings}
\label{apx:fine_tuning}
\begin{table*}[htbp]
\centering
\small
\begin{tabular}{cccccc}
\toprule
\textbf{Raw Input} & \textbf{Tokenized Input} & \textbf{Morphemes} & \textbf{POS (Morphemes)} & \textbf{POS (Segments)}  & \textbf{POS (Multitags)}       \\
\midrule
 & \multirow{2}{*}[-2pt]{\cjRL{b}}       & \cjRL{b}                  & \texttt{ADP}                      & \multirow{2}{*}[-2pt]{\texttt{ADP+DET}} & \multirow{2}{*}[-2pt]{\texttt{ADP+DET+NN}} \\
\cmidrule{3-4}
 &    & \cjRL{h}                    & \texttt{DET}                   &      &              \\
\cmidrule{2-6}
 & \cjRL{b}  &  &  & {\texttt{NN}} & \texttt{ADP+DET+NN} \\
\cmidrule{2-2}\cmidrule{5-6}
 & \cjRL{y}  & \cjRL{byt} & \texttt{NN}    & \texttt{NN} & \texttt{ADP+DET+NN} \\
\cmidrule{2-2}\cmidrule{5-6}
\multirow{2}{*}[-2pt]{\cjRL{bbyt hlbn}} & \cjRL{t}  &                      &                       & \texttt{NN} & \texttt{ADP+DET+NN} \\
\cmidrule{2-6}
 & \_ &                      &                       & \texttt{VOID} & \texttt{VOID}         \\
\cmidrule{2-6}
 & \cjRL{h}  & \cjRL{h}                    & \texttt{DET}                   & \texttt{DET}  & \texttt{DET+ADJ}      \\
\cmidrule{2-6}
 & \cjRL{l}  & &  & \texttt{ADJ}  & \texttt{DET+ADJ}      \\
\cmidrule{2-2}\cmidrule{5-6}
 & \cjRL{b}  & \cjRL{lbn} & \texttt{ADJ} & \texttt{ADJ}  & \texttt{DET+ADJ}      \\
\cmidrule{2-2}\cmidrule{5-6}
 & \cjRL{n}  &                      &                       & \texttt{ADJ}  & \texttt{DET+ADJ}    \\ 
\bottomrule
\end{tabular}
\caption{Input and output formats for fine-tuning. Whitespaces are assigned with the \texttt{VOID} tag.}
\label{tab:tagging-formats}
\end{table*}

We consider two mappings from morphemes to characters: \textit{multitags}, and \textit{segments}.
Table~\ref{tab:tagging-formats} illustrates each mapping on the Hebrew example \cjRL{bbyt hlbn} (\textit{in the White House}).

\paragraph{Multitag}
This mapping assigns a single label for each word: the \textit{set} of its constituent morphemes' tags. 
For example, the word \cjRL{hlbn} comprises two explicit morphemes, \cjRL{lbn}+\cjRL{h}, where \cjRL{h} (\textit{the}) is a determiner and \cjRL{lbn} (\textit{white}) is an adjective.
With multitags, the entire word will be labeled as \texttt{DET+ADJ}, which is treated as a single class.
We then copy this label across each character in the word.

\paragraph{Segments}
For a higher-resolution mapping, we assign each character the label of its encompassing morpheme.
Due to phonemic mergers, some characters take part in more than one morpheme, resulting in character-level multitags.
For example, the word \cjRL{bbyt} is composed of the morphemes \cjRL{byt}+\cjRL{h}+\cjRL{b} (\textit{in}+\textit{the}+\textit{house}), where the middle morpheme (\cjRL{h}) is covert, thus its POS tag is appended to that of the previous overt morpheme \cjRL{b}.

\end{document}